\documentclass[letterpaper, 10 pt, conference]{ieeeconf}  

\IEEEoverridecommandlockouts                              

\usepackage{graphics} 
\usepackage{cite}
\usepackage{amsmath,amssymb,amsfonts}
\usepackage{algorithmic}
\usepackage{graphicx}
\usepackage{textcomp}
\usepackage{xcolor}

\title{\LARGE \bf
Titan: A Parallel Asynchronous Library for Multi-Agent and Soft-Body Robotics using NVIDIA CUDA}

\author{Jacob Austin$^{1}$, Rafael Corrales-Fatou$^{2}$, Sofia Wyetzner$^{1}$, Hod Lipson$^{1}$
\thanks{*This work has been funded in part by U.S. Defense Advanced Research Project Agency (DARPA) TRADES grant number HR0011-17-2-0014}
\thanks{$^{1}$Department of Computer Science, \textit{Columbia University}, New York, NY. Correspondences to \textit{jacob.austin@columbia.edu}}%
\thanks{$^{2}$Imperial College London, W2 1PF London, UK, \textit{rafael.corrales-fatou19@imperial.ac.uk}}}
\begin{document}

\maketitle
\thispagestyle{empty}
\pagestyle{empty}

\begin{abstract}

While most robotics simulation libraries are built for low-dimensional and intrinsically serial tasks, soft-body and multi-agent robotics have created a demand for simulation environments that can model many interacting bodies in parallel. Despite the increasing interest in these fields, no existing simulation library addresses the challenge of providing a unified, highly-parallelized, GPU-accelerated interface for simulating large robotic systems. Titan is a versatile CUDA-based C++ robotics simulation library that employs a novel asynchronous computing model for GPU-accelerated simulations of robotics primitives. The innovative GPU architecture design permits simultaneous optimization and control on the CPU while the GPU runs asynchronously, enabling rapid topology optimization and reinforcement learning iterations. Kinematics are solved with a massively parallel integration scheme that incorporates constraints and environmental forces. We report dramatically improved performance over CPU-based baselines, simulating as many as 300 million primitive updates per second, while allowing flexibility for a wide range of research applications. We present several applications of Titan to high-performance simulations of soft-body and multi-agent robots.

\end{abstract}

\section{INTRODUCTION}

Massively-parallel GPU computing has seen widespread adoption in many disciplines, including machine learning and graphics processing, but it has not been as widely used in robotics research and simulations. Robotics is often inherently serial and low-dimensional, leaving little room for robotics problems to benefit from the massively parallel computation afforded by GPUs. Soft-body and multi-agent robotics, on the other hand, present many opportunities for extreme parallelism in kinematics computations and dynamic optimization. Recent work in soft robotics has enabled the creation of flexible robots capable of complex motion and environmental adaptation without rigid mechanical components \cite{softbody1}. Other authors have imitated the remarkable locomotion of inchworms with complex actuation of flexible alloys\cite{worm3}. Yet these robots cannot be simulated without modeling the complex internal interactions of thousands of pliable structural components, a computationally expensive and algorithmically complex task perfectly suited to GPU parallelism.

Multi-agent autonomy and multi-agent robotics have also seen an explosion of interest. These robots are composed of ensembles of interacting agents and use their collective interactions to produce collective motion. Recent work has created actuating cellular robots capable of coherent group motion \cite{multiagent2} and insect inspired small robot swarms \cite{multiagent1}. While each robot may itself be simple, modeling many interacting agents is beyond the scope of most existing software products. Finally, reinforcement learning has been one of the most successful models for learning complex locomotion and control without human supervision. The ability to run multiple, parallel simulations is of the utmost important, and scaling to many simultaneous simulations can substantially improve learning rates. Many papers in these areas have been required to build custom software to address their simulation needs, including raw speed, asynchronicity between the simulation environment and a CPU-based optimization algorithm, and flexibility to construct a wide variety of robots. 

In this work, we present a GPU-accelerated C++ software library, \textbf{Titan}, written using NVIDIA CUDA \cite{cuda}, which uses a high-performance asynchronous simulation system to model the behavior of soft-body and multi-agent robots at massive scales. This library decomposes soft-body problems into a spring-mass system and can then simulate hundreds of thousands of masses and millions of springs in real time, achieving an update rate of 300 million spring updates per second on reasonable hardware, with arbitrary scaling to new generations of GPUs. This model is not only fast but flexible, allowing a user to run the simulation on the GPU while performing simultaneous analysis and optimization on the CPU. Some higher-level primitives like containers, actuating robots, constraints, and surfaces are included, and others can be easily added with inheritance and runtime polymorphism. 

The primary innovations of this library are its asynchronous computation model and its powerful CUDA kernel design for kinematics simulations in parallel, both essential for simulating and optimizing actuating soft robots in real time. The asynchronous computation model allows the CPU and GPU to operate separately and synchronize only when needed, keeping objects on the GPU whenever possible to minimize copying. This means a massive simulation can run while the CPU processes existing data on virtual objects, with the GPU pausing its simulation only long enough for those changes to be propagated to the GPU. A flexible breakpoint system freezes the simulation where desired and allows the modification of GPU objects without any GPU-CPU copying bottlenecks. Topology optimization simulations that add or remove internal components can do so in constant time in response to an optimization algorithm while the simulation is running. Multiple reinforcement learning agents can be simulated in parallel on an arbitrary number of GPUs. The CUDA kernel design is also very powerful, alternating updates of springs and masses to avoid race conditions. All springs apply forces to their corresponding masses at once, with either accumulation-based or atomic addition for multiple springs attached to a single mass; then a mass kernel runs applies constrained kinematics based on applied constraints, solid bodies, and friction. These alternating kernels do not have to synchronize and can update objects with different time resolutions. 

In this work, we describe the simulation libraries and associated GPU algorithms for achieving extremely large-scale simulations at high speeds and implement a variety of soft-body and multi-agent robots which are simulated at massive scale.

\section{RELATED WORK}

\subsection{GPU-Accelerated Robotics Libraries}

Due to the inherently serial nature of many robotics applications, few attempts have been made to develop GPU accelerated robotics simulation environments. The most popular existing GPU library for physics simulations is the NVIDIA FleX simulation library \cite{nvidiaflex}, a toolkit developed by NVIDIA with a GPU accelerated simulation environment. Primarily designed for visual effects, including cloth and fluid animations, there is limited support for real-time simulations with CPU-GPU communication in real time or fine-tuned resolution controls. While more fully featured in some respects, it sacrifices performance for very large objects which Titan is able to simulate easily. NVIDIA Flex is also not fully open-source. Other libraries, especially for computer vision applications, like Carla, use GPU acceleration primarily for environmental rendering and ray-casting, not for handling the physics of the robot itself \cite{carla}.

\subsection{Soft-Body and Multi-Agent Robotics Simulations}

Several libraries exist for simulating soft-body physics in general, most prominently the NVIDIA FleX and PhysX toolkits, are able to handle soft objects, but rely on a constraint-based physics solver that cannot accommodate many common mechanisms for soft-body actuation \cite{nvidiaflex}. Several authors (\hspace{1sp}\cite{etheredge}, \cite{mesit}) have developed algorithms for high-performance parallel spring-mass simulations on the GPU, using a method for interleaving spring and mass updates to achieve high performance from CUDA kernels. Coevoet et al. develops a simulation library for soft robots with a particular focus on medical applications \cite{coevoet}. Our work builds on this in some ways, but their library is not intended to achieve high-performance for large systems. Wang et al. also decomposes soft-body robots into a spring-mass system \cite{wang}. Hu et al. presents a unique new approach to modeling soft robotics with a differentiable simulator that can be incorporated into gradient-based optimization problems \cite{chainqueen}.

\subsection{Reinforcement Learning Simulation Environments}

Reinforcement learning libraries place a significant emphasis on parallelism and performance but have not widely used GPUs for the actual physics simulations. Libraries like MuJoCo \cite{mujoco} have many of the same requirements as Titan - high performance, asynchronicity, simultaneous optimization and simulation, and parallelism. MuJoCo does support parallel simulations on multiple CPU cores, but it does not use GPUs for parallel simulations. MuJoCo also uses Euler and Runge-Kutta integrators instead of a more sophisticated Hamiltonian solver, like Titan. DART and Bullet are also popular environments, but MuJoCo remains the most popular simulator in part due to its inclusion in the OpenAI lab environment. However, MuJoCo runs exclusively on the CPU and has struggled to port their library to the GPU with significant performance gains over the current multi-CPU configuration. Liang et al. at NVIDIA published exploratory work in GPU acceleration for RL simulations that achieved enormous improvements over a CPU-based algorithm \cite{liangnvidia} with an NVIDIA Flex-based software system capable of simulating human locomotion.

\section{GPU LIBRARY DESIGN}

The following sections describe the key GPU algorithms of the Titan library that facilitate parallel simulations and optimizations for robotics applications.

\begin{figure}[h!]
	\centering
	\includegraphics[scale=0.6]{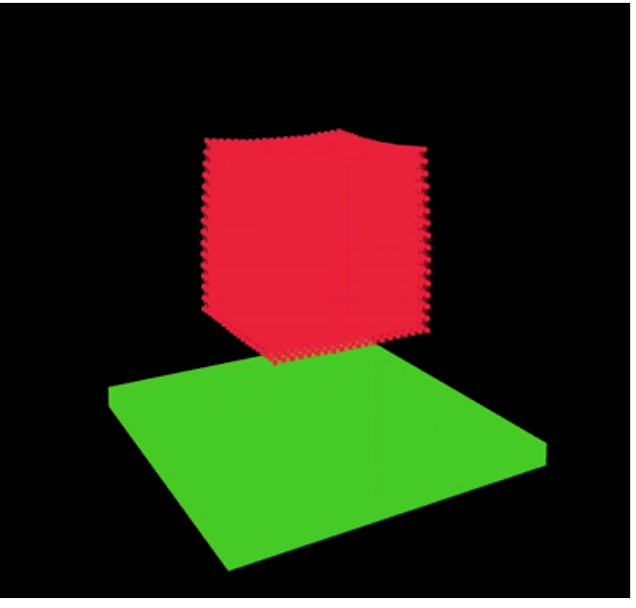}
	\caption{A large 30x30x30 spring-mass lattice in the Titan library}
\end{figure}

\subsection{Design Overview}

The Titan library is written in C++ using NVIDIA CUDA, encompassing a system for decomposing soft robotic structures into spring-mass lattices and a a parallel Euler iteration method for iteratively updating the dynamics of the soft-body structure. This abstraction captures all of the requirements needed to discretize and simulate soft robots or even solids with some degree of flexibility. Masses are updated with a specified time resolution and constrained by a set of possible constraints, including plane and line constraints, subject to frictional and atmospheric forces. Solid objects like planes and spheres can be created in the environment to constrain objects further and apply atmospheric and frictional forces.

\begin{figure}[h!]
	\centering
	\includegraphics[scale=0.28]{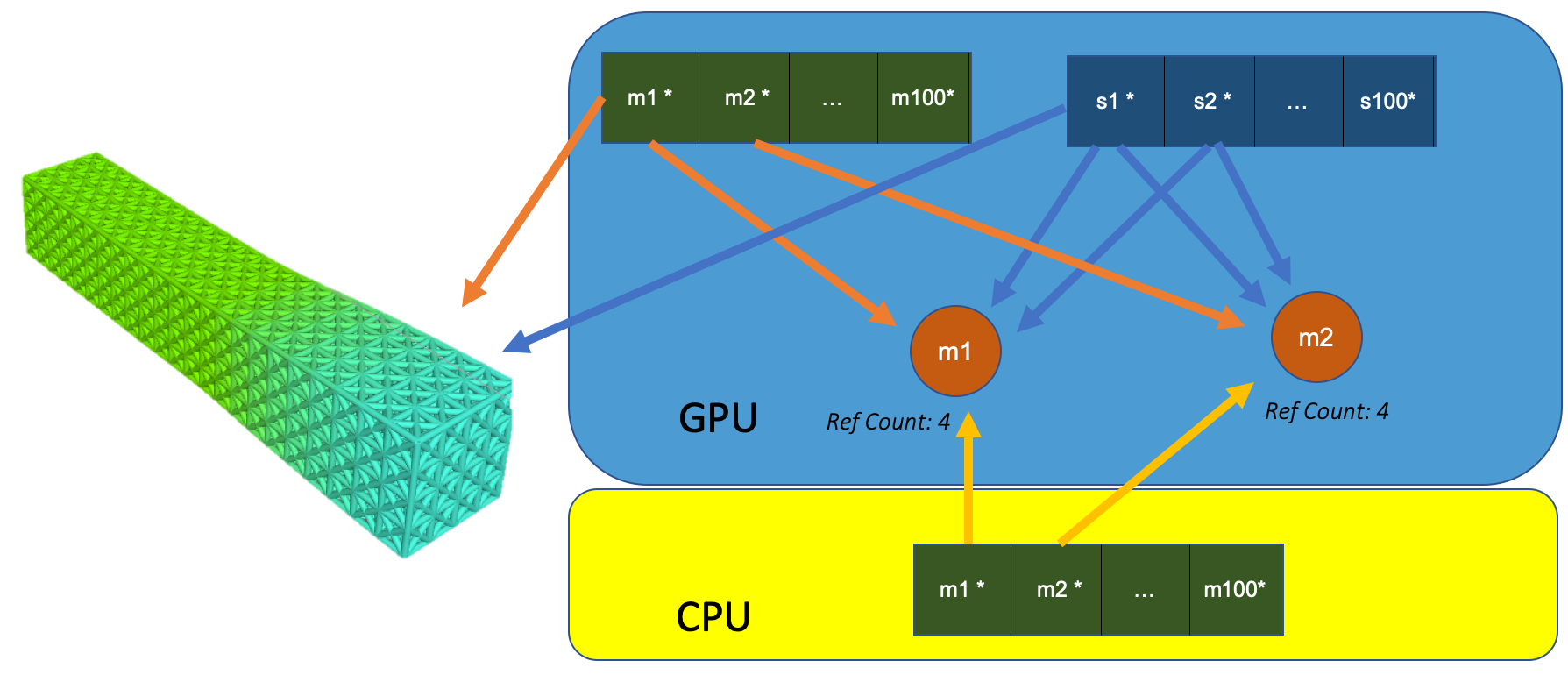}
	\caption{Left: worm robot simulated in Titan. Right: architecture design of fast data structures, with reference counting for GPU and CPU objects.}
\end{figure}

\subsection{CUDA Kernel Design for Fast Spring Mass Simulations}

All physics simulations in Titan operate using an iterative explicit Euler kinematics solver, applying forces using Hooke's Law and propagating the resulting kinematics at a fixed resolution. While many modern simulation libraries use more sophisticated kinematics solvers with higher order terms, the raw speed of parallelism on the GPU makes a simpler Euler, Verlet, or RK4 integration scheme with small time-steps a viable alternative to more expensive and more serial solvers \cite{mujocobook}. The ability to alternate applying Hooke's law forces and kinematics solvers inspires a natural iterative approach to simulating physics in parallel, first iterating over all springs and applying a Hooke's Law force to each corresponding mass, and then iterating over all masses to update their accelerations and positions using, in the simplest case,
 
\begin{equation}\mathbf{F}=k(x_1 - x_2)\qquad\text{and}\qquad\Delta x = \dot{x}\Delta t + \ddot{x} (\Delta t)^2\end{equation}

The mass updates are totally independent and can be implemented as a single CUDA kernel with a GPU thread for each mass. The spring updates are generally independent except where multiple springs are connected to a single mass. Titan handles this in two ways. The easiest approach is to enforce serial addition using a mutex lock on each mass or an atomic addition operation. Recent NVIDIA GPUs all support atomic vector addition, and usually simultaneous access is rare enough to make these approaches computationally tractable. In experiments, these approaches incurred no more than a 1.25x performance loss over more sophisticated methods. The second approach is an accumulation-based method that allocates an array of force vectors for each mass. Each spring connected to a mass is assigned a single target location in the vector to store the force it applies, and after the full update, the mass iteration can sum all of these forces using another CUDA kernel and apply the force. This is more difficult to implement and incurs a memory overhead for storing these accumulation arrays. This also requires masses to keep track of the springs attached to them and to enforce an ordering over the attached springs. For this reason, atomics are the default choice despite the advantages of this approach.

\subsection{Fast Data Structures for O(1) Insertion and Deletion}

Any robotics simulation that frequently modifies the topology of the robot requires the ability to easily add and remove beams, joints, and springs without significant overhead. For this application, we designed an array data structure on the GPU that uses lazy deletion and indirection to achieve constant-time insertion and deletion of masses and springs on the GPU.

All robotics components reside on the GPU unless explicitly copied to the CPU. Pointers to \textit{Spring} and \textit{Mass} objects are stored in an array on the GPU, and the objects themselves are generally stored in arrays as well, although they can be stored elsewhere if created separately. This indirection model allows individual masses and springs to be created and deleted lazily in O(1) time, simply by marking the pointers as NULL and referencing counting the objects themselves. 

Both GPU and CPU references are counted, so an object can be analyzed in simulations even after it has been destroyed. Arrays of pointers also allow amortized O(1) insertion, since they are made larger than the number of masses required, and can be extended like linked lists. Global constraints are also stored in a linked list provided by the NVIDIA Thrust library, while local constraints can be stored in global memory and referenced via a pointer stored in the corresponding mass object. Masses do not keep track of the springs connected to them, so there is no additional book-keeping for adding or deleting springs. 

This capability is often used for topology optimization of soft-body robots or 3D printing, simulating a mesh under external stress, adding or removing internal supports as needed to attain a low-cost, high-strength body. 

\subsection{Asynchronous CPU/GPU Computation Model}

Likewise, any system that performs real-time optimization or topology modification, like reinforcement learning models, must be able to optimize the robotic structure based on results from the simulation. To this end, Titan has developed a fully asynchronous communication model between the CPU and GPU. A thread launched on the CPU constantly launches CUDA kernels for the spring and mass updates, which waits only for breakpoints created by any of the main threads. Operations are thread and process safe by design. Once the simulation is started, updates occur as quickly as possible, without copying between the CPU and GPU. At a specified time, with the \textit{sim.setBreakpoint()} command, or on a specific simulation condition, the GPU will halt execution and wait for any CPU thread to reactivate it. The CPU thread can call \textit{sim.waitForEvent()} to block until a breakpoint has been reached when no other computations are desired.

\begin{figure}[h!]
	\centering
	\includegraphics[scale=0.3]{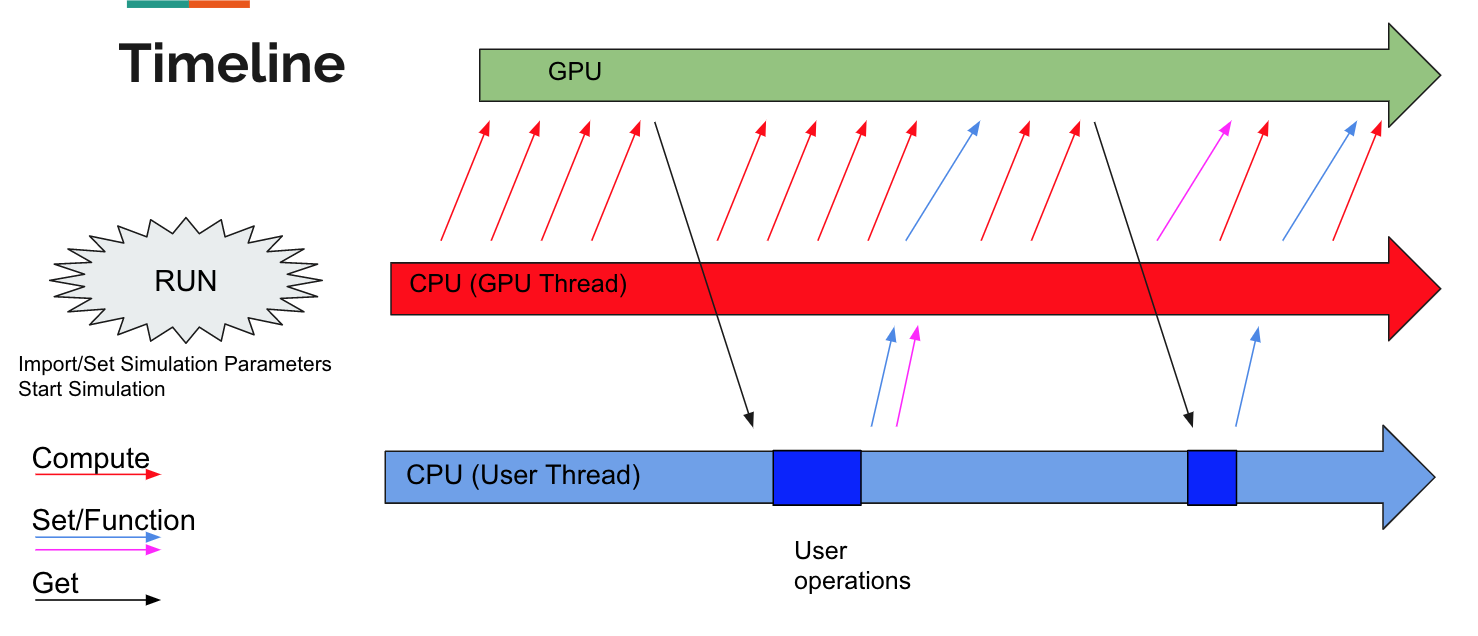}
	\caption{Asynchronous Computation Model}
	\label{async}
\end{figure}

Figure \ref{async} shows this computation model. The GPU thread repeatedly launches CUDA kernels (bypassing the kernel execution limit), while the CPU operates freely and only blocks when it deliberately issues a blocking command. Copying to and from the GPU can only occur when the GPU simulation has stopped, but virtual objects on the CPU can be modified while the GPU is running and pushed to the GPU quickly with minimal interruption. Updates can be queued and pushed to the GPU in a small period of time where the simulation is paused, for example after a reinforcement learning or evolutionary algorithms epoch.

\subsection{Simulating Real Materials}

Trusses and solids approximations made of real materials can be simulated by deriving spring constants and densities from real parameters. One may convert an elasticity modulus into corresponding spring constants using the following formula:

\begin{equation}
k = EA_{c}/L_{r}
\end{equation}

Where $E$ is the elasticity modulus of the material, $A_{c}$ is the cross-area of the spring (Titan optionally stores spring diameters on the data structure), and $L_{r}$ is the rest length of the spring. For instance, a simulation of a nylon with and elasticity modulus of 4.56 GPa using a component spring of length 1 cm and diameter of 1 mm would produce a spring constant of $3.58e5$.

Individual mass values can be derived from the material density and from either the volumetric tetrahedra underlying the mesh structure in the case of a solid object approximation, or from the spring measurements in the case of a truss simulation where each spring is treated as a bar and half of each bar's mass is summed on the connecting mass node.

In addition, material yield values can be added to be compared against in force calculation to determine "broken" springs. Volumetric actuation will be discussed in later examples, and it also achieved by setting material parameters. Note that actuation is considered abstracted from the method of actuation, therefore only volume and frequency parameters are specified directly instead of environmental triggers such as temperature etc.

\subsection{Soft-Body Lattice Construction}

The Titan library is designed to support a variety of soft robots and flexible solids with an internal lattice structure. To this end, the library has several novel construction algorithms for filling a defined solid boundary with a flexible lattice of beams and joints. The library can import shapes from standard STL files and fill standard shapes like cubes, rectangles, and spheres with a spring/mass lattice. This is done using a ray-casting approach often used in STL rendering. The ability to import complex objects and construct a fine mesh within them using real materials allows Titan to function in a wide number of scenarios for any robotic shape or layout.

\section{EXPERIMENTS}

Many experiments were performed to validate the physics of the Titan library and compare its performance to other GPU and CPU implementations. All experiments were performed on an NVIDIA Titan X GPU and a 3.7GHz Intel Core i7-8700K CPU.

\subsection{Bouncing Cubes}

\begin{figure}[h!]
	\centering
	\includegraphics[scale=0.35]{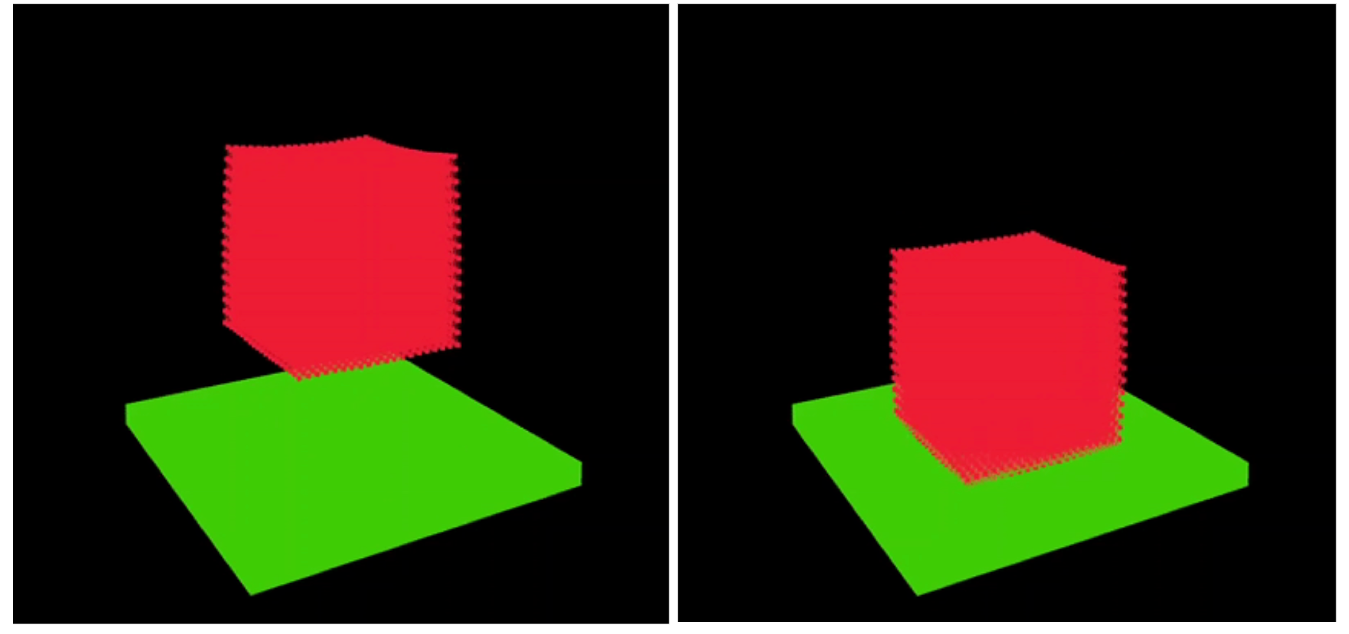}
	\caption{28x28x28 mass lattice with 15000 springs. \textit{Left}: starting position, \textit{right}: position at 2 seconds}
	\label{lattices}
\end{figure}

To validate the physics and performance of the GPU library, we benchmarked the performance on a system of large spring-mass lattices with tens of thousands of masses and hundreds of thousands of springs at various integration timesteps. An example of this lattice is shown in Figure \ref{lattices}.

\begin{figure}[h!]
	\centering
	\includegraphics[scale=0.35]{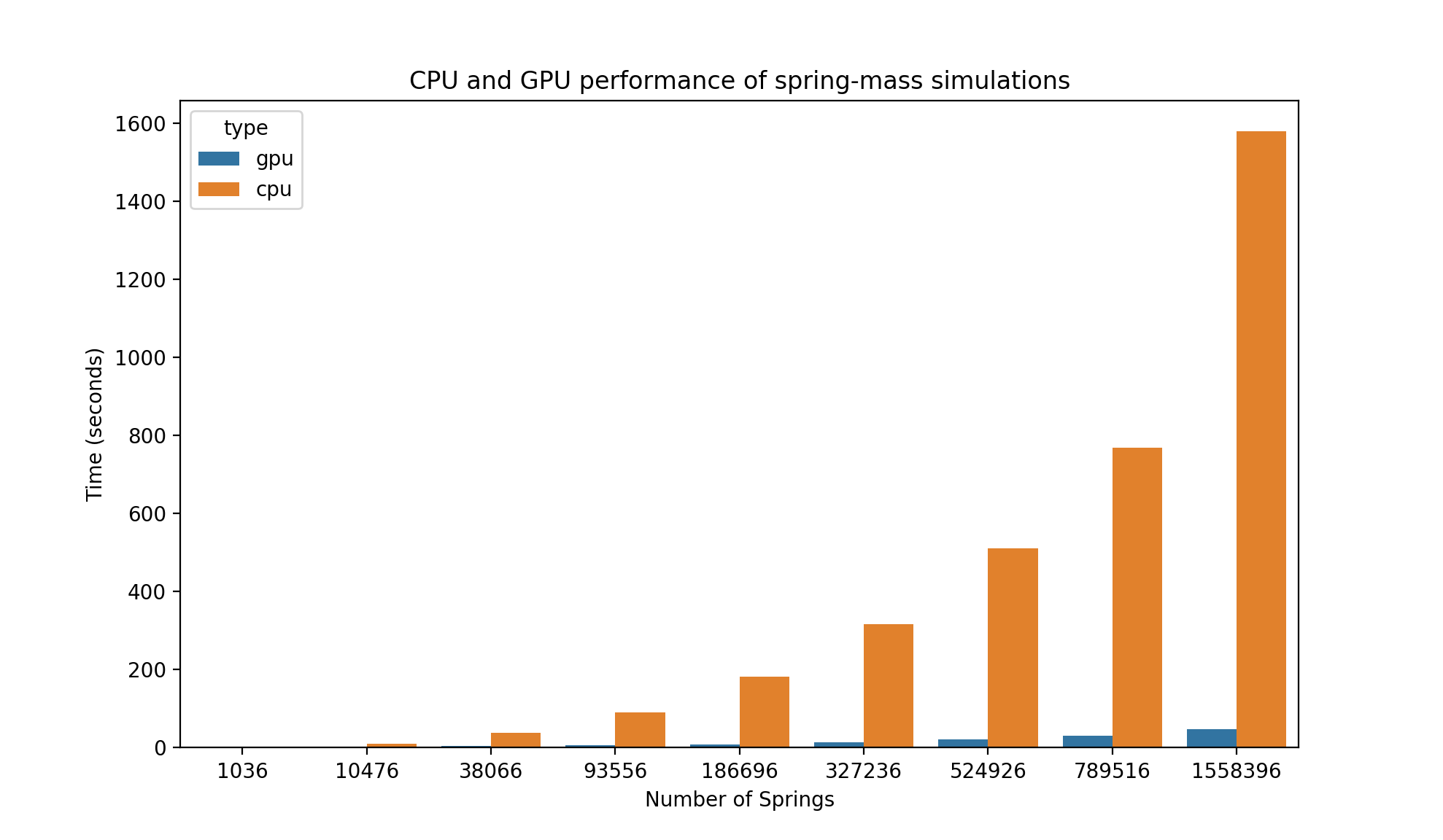}
	\caption{Runtime for 1 second simulation of spring-mass lattices with 0.0001 temporal resolution on CPU and GPU.}
	\label{performance}
\end{figure}

These lattices are built from locally interacting springs, and can be used as basic units for constructing more complex solids. Our benchmark involved simulating this lattice system for a 1-second period with a time-step of 0.0001, on the final GPU version and an identical CPU version that uses a sequential loop to apply force and position updates. The performance results for the lattice simulation with different lattice densities is shown in Figure \ref{performance}. Performance is roughly equal for very small lattices, but CPU performance scales linearly with the number of springs in the system, while the runtime on the GPU remains roughly constant, even for extremely large lattices. For example, a 50x50x50 mass lattice with 1,558,396 springs running with a delta-t of 0.0001 for a one second simulation using the Titan library takes 40.01524 seconds. This includes $10000$ unique updates for a rate of $1558396 * 10000 / 40.01524$ updates per second, a total of $389450619$ spring updates per second. This figure can be improved further on new hardware.

The CPU takes 1573.24 seconds to run the same simulation serially, a 3900\% performance increase. 

\subsection{Soft Robotics Application: Locomotive Worm Simulation}

An important application of this soft-body simulation capability is soft-body locomotion achieved from thousands of interacting components. There are many fabricated examples of soft robotics worms (\hspace{1sp}\cite{worm1}, \cite{worm2}, \cite{worm3}) that exhibit peristaltic actuation patterns in order to move. These are based on the locomotive mechanics found in real earthworms, which have a segmented structure where each segment expands or contracts its muscles in turn to achieve movement. Our simulated worm is embodied by a rectangular cubic lattice that is actuated to produce waves of motion. We posit a material that has coordinated actuation such that each horizontal section on the worm has a periodic expansion that is offset from its neighbors respectively. The actuation is applied during integration with the following formulas:

\begin{equation}
t = (T - t_{o}) \bmod t_{p}	
\end{equation}
\begin{equation}
L_{s} = (1 + c * \sin(\omega * t))
\end{equation}

Where $t$ is the repeated point in the periodic time scale, $T$ is the simulation time, $t_{o}$ is the actuation offset, $t_{p}$ is the actuation period, $c$ is the expansion/contraction factor, and $\omega$ is the frequency constant.

\begin{figure}[h!]
	\centering
	\includegraphics[scale=0.25]{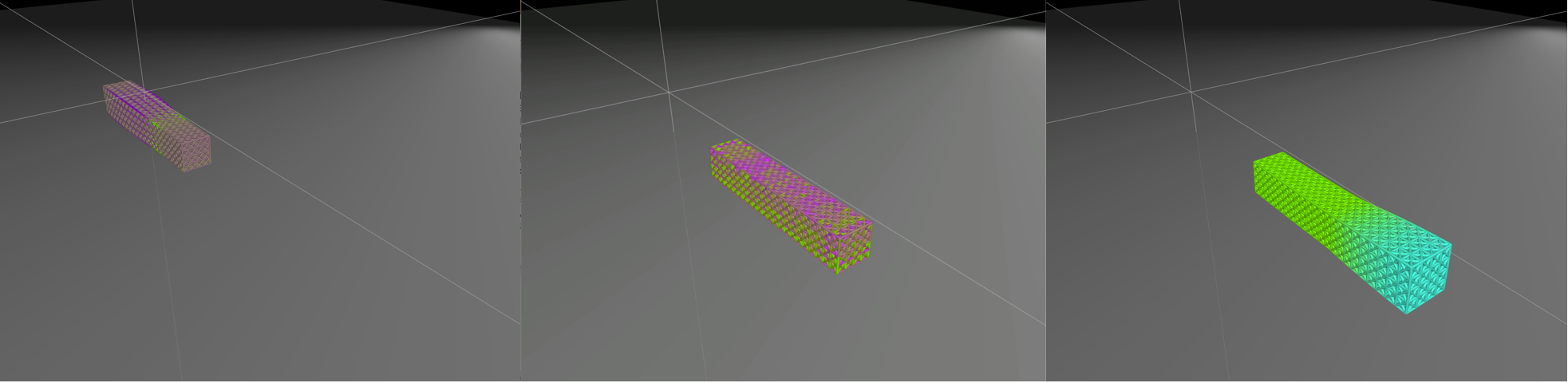}
	\caption{Simulated locomotive worm at timestep 7200, 51000, and 138600 respectively. (a) and (b) show the current forces applied by each spring, where purple indicates expansion and green, contraction. (a) Shows the forces caused by actuation, (b) shows relaxation of the structure. (c) Visualizes the actuation factor itself.}
	\label{worms}
\end{figure}

Illustrated in [Fig. \ref{worms}], the parameters used for this example include

\begin{equation}
t_{o} = \min(x_{m_{1},0}, x_{m_{2},0}) - x_{min,0}
\end{equation}

where $m_1$ and $m_2$ are the connected, $x_0$ refers to the starting x-component of their position vectors, and $x_min$ is the minimum x-position represented in the lattice, which will be the starting point of the actuation wave. Other parameters were set to an actuation period of $t_p = 1.0 s$, a frequency of $\omega = 20$, and an expansion constant of $c = 0.2$; These produce a periodic wave of out-of-sync actuations across the lattice, that pause fully before resuming. We found that this pattern creates a satisfactory locomotion as shown in [\ref{worms}].	

We were also able to simulate hundreds of these robots, with a total of 339200 springs in real time. 

\begin{figure}[h!]
	\centering
	\includegraphics[scale=0.2]{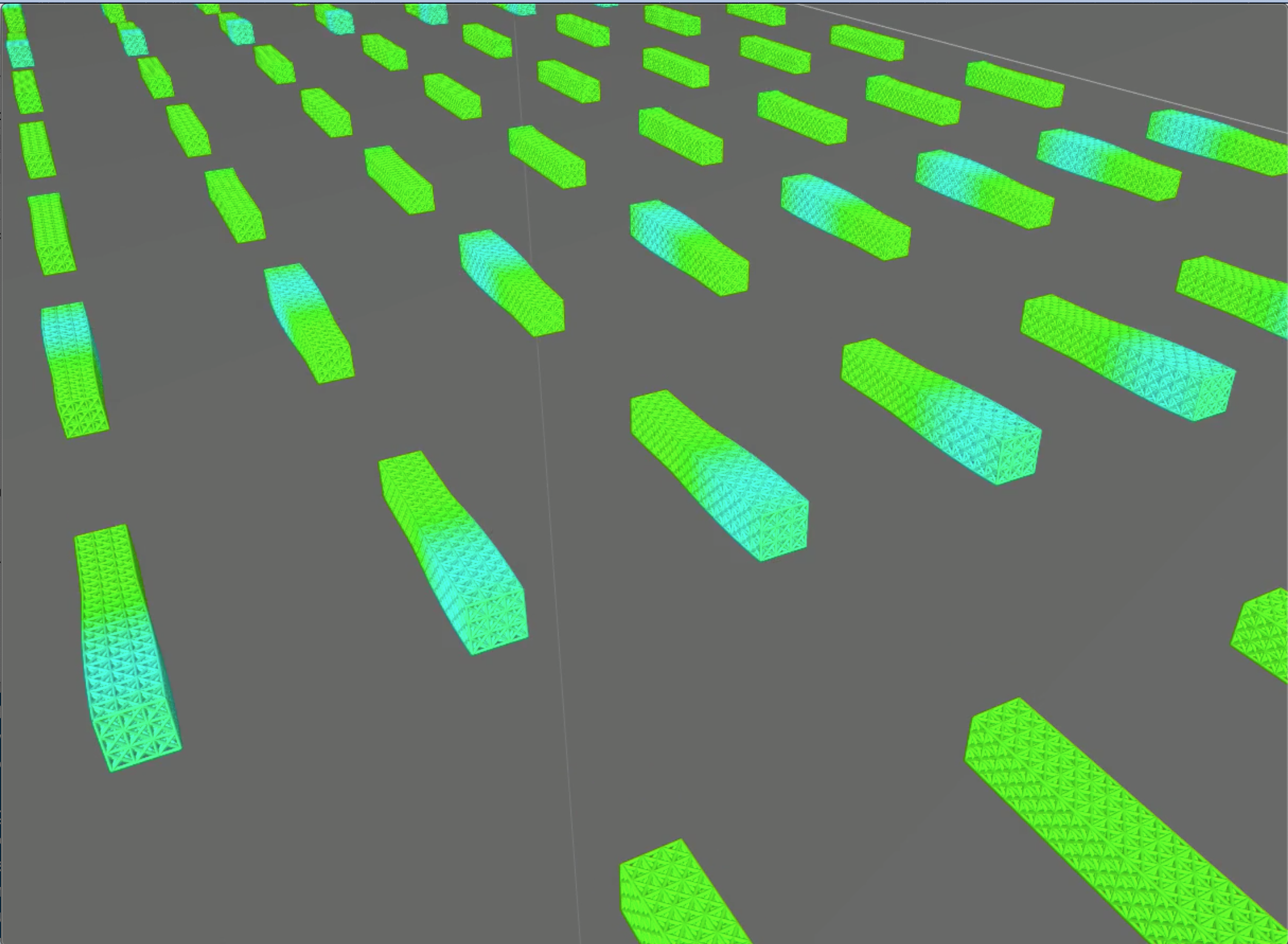}
	\caption{100 locomotive worm robots simulated in parallel using Titan.}
	\label{manyworms}
\end{figure}

\subsection{Soft Robotics Application: Multi-Body Swarms}
	
Many recent papers have introduced new techniques in multi-agent robotics, but many have been forced to build a custom simulation environment for each new project (\hspace{1sp}\cite{multiagent1}, \cite{multiagent2}). Titan presents a more flexible and universal option for simulating these systems. Titan allows for a variety of robotic systems to be abstracted into an approximate form that is fast, featured, and extensible. In this vein, we have created a proof-of-concept simulation of loosely-coupled actuated bodies in order to demonstrate the potential for Titan to fill this space. 

\begin{figure}[h!]
	\centering
	\includegraphics[scale=0.25]{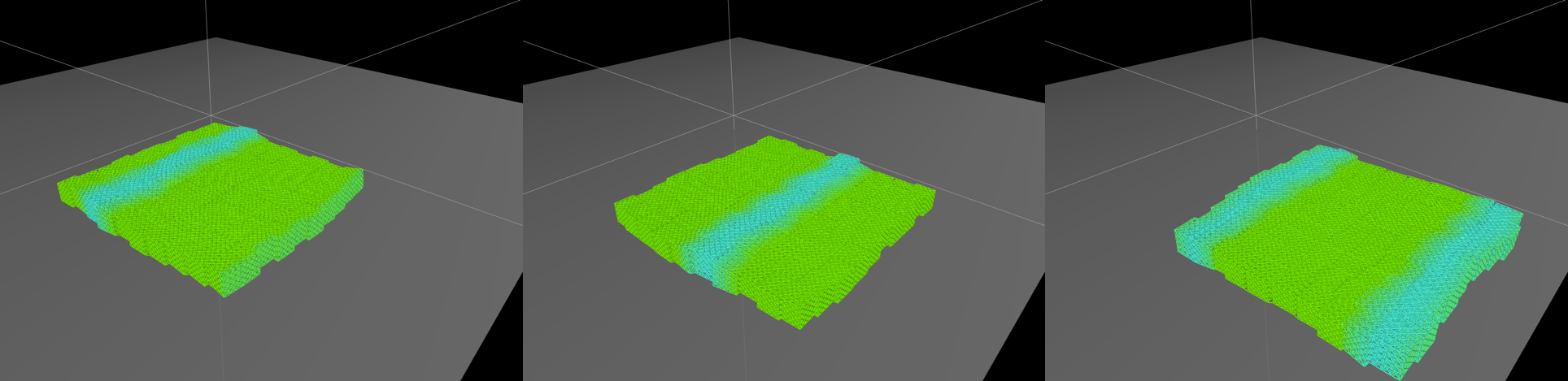}
	\caption{Simulated multi-body robot at timesteps 12000, 178800, and 381000 respectively. Blue indicates expansion is applied through actuation.}
	\label{multibody}
\end{figure}

Springs here approximate solid materials as well as connective magnetic forces, which use a much lower spring constant to suggest a weaker connection. The bodies are actuated according to the same scheme as the worm example, i.e. as a function of time and starting position. As each body actuates, the group mass begins to locomote ()Figure \ref{multibody}). As pictured, 100 cubes are loosely connected with thin horizontal planes of springs at their tops and bottoms. These springs have diameters of 0.4 mm, which creates an artificially small spring constant, compared to the 1.0 mm diameters of the springs within each cube. As the group moves, the springs all follow the same actuation pattern. The connective springs transfer forces between cubes weakly, but enough to produce a coordinated motion. Note that to reflect how expansion and contraction might be achieved physically by a variety of different methods, the actuation pattern is entirely definable and at all not limited to the method shown here. This robot includes 100000 masses and 784234 springs, and was constructed with just 20 lines of C++ code using the library. A custom solution implementing this interacting, locomotive system would likely require thousands of lines of code without using Titan.

\subsection{Topology Optimization}

The asynchronous communication with the GPU allows for not only dynamic updates to multiple bodies, but also for creation, deletion, and updating of the components within bodies during runtime. This has facilitated rapid improvement in topology optimization projects for objects under dynamic forces, along with actuation and shape optimization applications. Titan enables deforming the underlying mass-spring mesh during simulation time by updating masses, spring constants, spring rest lengths, etc., on the CPU and pushing to the GPU at controlled intervals. For topology optimization, springs might be dynamically added or removed under certain stress thresholds. For shape optimization, the user can remove and "grow" mass from an existing object in simulation. For actuation, springs might expand and contract as a function of position and time, with the CPU calculating this function and pushing the result along. These applications all exemplify successful internal applications of Titan and reflect the flexibility of asynchronously combining simulation and control.

\section{FUTURE WORK}

We see great of potential for Titan as a platform to expand upon. Our lab has already explored avenues in using Titan as a base engine to model robotic components, and to perform topology optimization. New CUDA kernels might be added to Titan for other time integration methods that could increase accuracy. There are several physical features such as collision detection that could be added to make Titan more robust as well. We are excited that the nature of Titan as an Open Source project will allow researchers to modify and improve its components as necessary. We provide a Python API to make the Titan library accessible to the robotics community.

Using Titan’s current features, there are many interesting systems that might be modeled. Objects can be in motion and then have their components altered in the midst of movement. Taken as a plausible physical phenomenon, this dynamic ”sculpting” opens up many possibilities from simulating adaptive materials to programmable matter.

\section{CONCLUSION}

In this paper, we introduced the Titan simulation library capable of large scale simulations of novel soft robots and multi-agent robotic systems. Novel algorithmic elements of the library permit extremely high-performance simulations and flexible construction of various kinds of robots. The asynchronous computation system makes simultaneous simulation and optimization convenient and flexible. Several examples of robot simulations were given that demonstrate aspects of the library design.

\addtolength{\textheight}{-12cm} 

\section{APPENDIX}

\subsection{Code}

Code is available at \textit{https://github.com/ja3067/Titan}.

\subsection{Python Bindings}

Titan includes a set of Python bindings which support all Titan operations except rendering using the OpenGL API. These bindings use Numpy arrays to hold positional information instead of a custom Vector3D class, but perform equally well since they make calls to the C++ backend.

\end{document}